\pdfoutput=1

\documentclass[11pt]{article}

\usepackage{amssymb}
\usepackage{amsfonts}
\usepackage{graphicx}
\usepackage{stfloats}
\usepackage{amsmath}
\usepackage{subfigure}
\usepackage{multirow}
\usepackage{bbm}
\usepackage[colorlinks, linkcolor=blue, anchorcolor=blue, citecolor=blue]{hyperref}

\usepackage{emnlp2021}

\usepackage{times}
\usepackage{latexsym}

\usepackage[T1]{fontenc}

\usepackage[utf8]{inputenc}

\usepackage{microtype}
\usepackage{url}
\usepackage{booktabs}       
\usepackage{amsfonts}       
\usepackage{amsmath,amssymb,amsthm}
\usepackage{algorithm}
\usepackage{algorithmic}
\usepackage{times}
\usepackage{soul}
\usepackage{url}
\usepackage{graphicx}
\usepackage{tabularx}
\usepackage{multirow}
\usepackage{subfigure}
\usepackage{mathrsfs}
\usepackage{bm}

\title{An Explicit-Joint and Supervised-Contrastive Learning Framework for Few-Shot Intent Classification and Slot Filling}

\author{Han Liu$^{1,2}$ \quad Feng Zhang$^{1,2,3}$ \quad Xiaotong Zhang$^{1,2}$\Thanks{~ Corresponding author.}  \\
{\bf \quad Siyang Zhao$^{1,2}$ \quad Xianchao Zhang$^{1,2}$} \\
School of Software, Dalian University of Technology$^1$ \\
Key Laboratory for Ubiquitous Network and Service Software of Liaoning Province$^2$ \\
School of Electronics Engineering and Computer Science, Peking University$^3$ \\
{\tt \{hanliu,zhangxt\}@dlut.edu.cn, zhangfeng@stu.pku.edu.cn} \\
{\tt zhao\_siyang@mail.dlut.edu.cn, xczhang@dlut.edu.cn}}

\begin{document}
\maketitle
\begin{abstract}
Intent classification (IC) and slot filling (SF) are critical building blocks in task-oriented dialogue systems. These two tasks are closely-related and can flourish each other. Since only a few utterances can be utilized for identifying fast-emerging new intents and slots, data scarcity issue often occurs when implementing IC and SF. However, few IC/SF models perform well when the number of training samples per class is quite small. In this paper, we propose a novel explicit-joint and supervised-contrastive learning framework for few-shot intent classification and slot filling. Its highlights are as follows. (i) The model extracts intent and slot representations via bidirectional interactions, and extends prototypical network to achieve explicit-joint learning, which guarantees that IC and SF tasks can mutually reinforce each other. (ii) The model integrates with supervised contrastive learning, which ensures that samples from same class are pulled together and samples from different classes are pushed apart. In addition, the model follows a not common but practical way to construct the episode, which gets rid of the traditional setting with fixed way and shot, and allows for unbalanced datasets. Extensive experiments on three public datasets show that our model can achieve promising performance.
\end{abstract}

\section{Introduction}
With the vigorous development of conversational AI, task-oriented dialogue systems have been widely-used in many applications, e.g., virtual personal assistants like Apple Siri and Google Assistant, and chatbots deployed in various domains \cite{DBLP:conf/emnlp/LiuZFFLWL19,DBLP:conf/acl/YanFLLZWL20}. Intent classification (IC) and slot filling (SF) are key components in task-oriented dialogue systems, and their performance will directly affect the downstream dialogue management and natural language generation tasks \cite{intent-cnn}. Traditional IC/SF models have achieved impressive performance \cite{sigdial/GuptaHK19}, but they often require large amount of labeled instances per class, which is expensive and unachievable in industry especially in the initial phase of a dialogue system.

Few-shot learning aims to solve the data scarcity issue, which can recognize novel categories effectively with only a handful of labeled samples by leveraging the prior knowledge learned from previous categories. Most few-shot learning studies concentrate on computer vision domain \cite{fei2006one,finn2017model,jung2020few}. Recently, to handle various new or unacquainted intents popped up quickly from different domains, some few-shot IC/SF models are proposed \cite{DBLP:conf/acl/GengLLSZ20,DBLP:conf/acl/HouCLZLLL20}. Nevertheless, these methods usually focus on a single task and do not attempt to address these two tasks simultaneously.

\begin{figure*}[t]
	\centering
	\includegraphics[height=8.5cm, width=15cm]{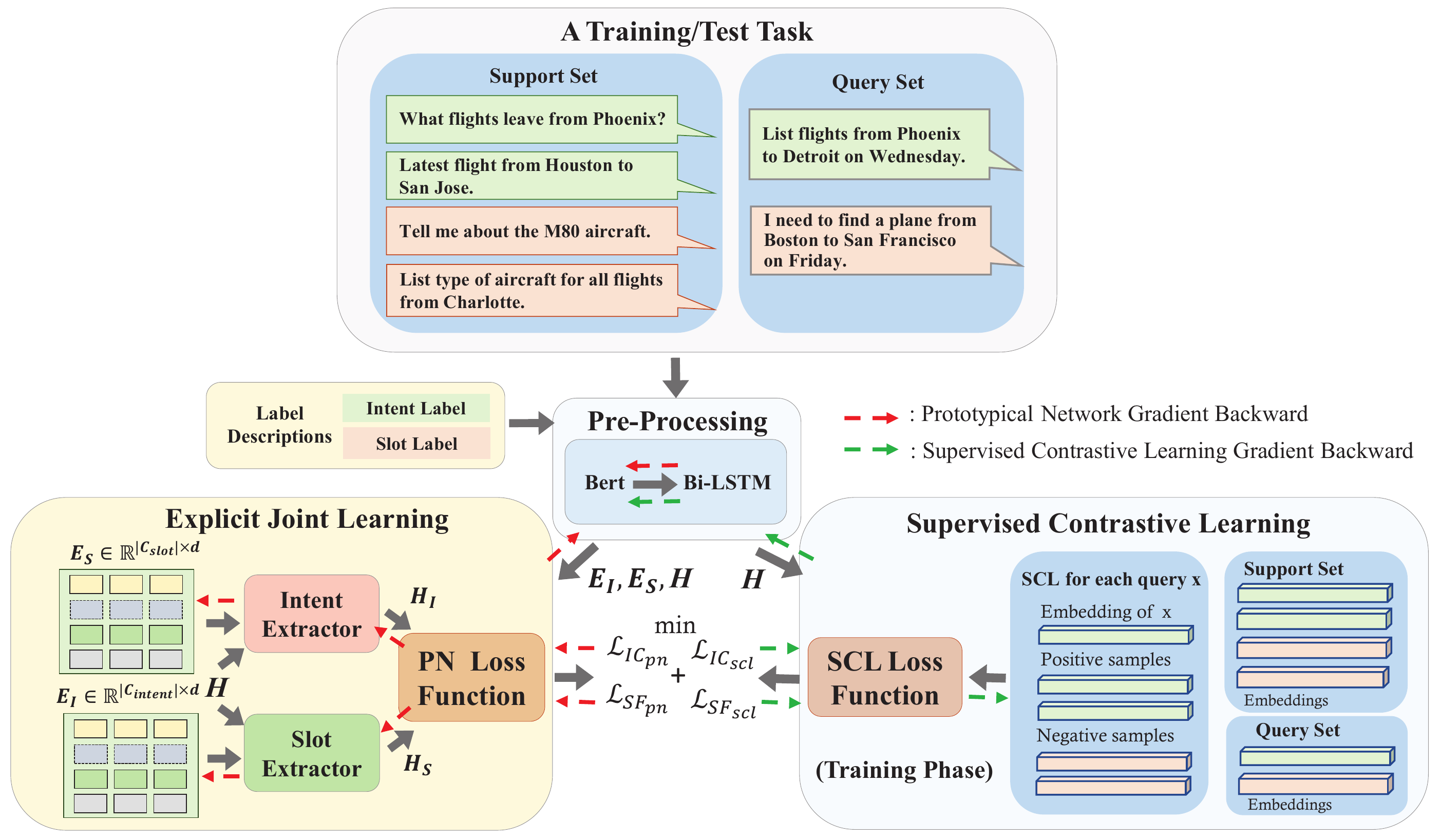}
	\caption{Illustration of our framework. In the training process, labeled utterances from support set and query set are first encoded by pre-processing module. Meanwhile, intent and slot labels' descriptions are fed into pre-processing module to generate intent embedding matrix and slot embedding matrix. Then the two matrices and utterance's embedding are fed into explicit joint learning module, while utterance's embedding is put forward into supervised contrastive learning module. In explicit joint learning module, intent and slot extractors are used to extract intent and slot information, which leverage the attention mechanism. Then, we can obtain slot-attention-based intent representation and intent-attention-based slot representation. Next, prototypical network uses intent labels to guide slot embedding learning and vice versa. In supervised contrastive learning module, we construct contrastive samples for each query instance using support set. And the SCL loss function can push samples from the same class more close and samples from different classes further apart. In the testing process, prototypical network is used to predict intent and slot labels, while supervised contrastive learning module is disabled. }
	\label{Fig.1}
\end{figure*}

Intuitively, IC and SF are two complementary tasks and the information of one task can be utilized in the other task to improve the performance. Existing joint IC and SF models have achieved impressive performance in supervised learning scenarios \cite{DBLP:journals/corr/abs-2101-08091}. But only a couple of methods are custom-designed for few-shot joint IC and SF task. \citet{handful-icsf} directly apply the popular few-shot learning models MAML and prototypical network to explore the few-shot joint IC and SF. During the same period, \citet{bhathiya2020meta} also attempt to utilize MAML to deal with this problem in a similar way. Though these models outperform the single task model, they just implicitly model the relationship between IC and SF. The mutual interaction between IC and SF in these methods is still unknowable, which seems to be a black box (not using a concrete formula to characterize the interaction), thus difficult to analyze the internal mechanism.

In this paper, we propose to model the relationship between IC and SF precisely and clearly, as well as integrating with contrastive learning. As illustrated in Figure \ref{Fig.1}, our framework consists of two main components. First, we present an explicit-joint learning framework for few-shot intent classification and slot filling, which effectively utilizes the bidirectional connection between IC and SF via leveraging slot-attention-based intent representation and intent-attention-based slot representation. In addition, we integrate with supervised contrastive learning to obtain more class-discriminative embeddings, which is a strong complementary part to improve our framework. 

To verify the effectiveness of the proposed model, we conduct extensive experiments on three public datasets. Catering to the unbalanced datasets and very limited labeled samples in real application scenarios, we adopt a not common but practical way to construct the episode for few-shot learning, i.e., in each episode, the way and shot are variable. The empirical study validates our proposal and shows promising results of our framework on IC and SF tasks.

\section{Related Work}
\paragraph{Few-shot learning}
Few-shot learning aims to use the knowledge learned from seen classes, of which abundant labeled samples are available for training, to recognize unseen classes, of which limited labeled samples are provided~\cite{few-shot-survey}. It has been widely studied in computer vision such as classification~\cite{fei2006one, classification-WangXLZF20},  segmentation~\cite{segment-WangLZZF19, corr/abs-1806-07373} and generation~\cite{iccv/0001HMKALK19}. Recently it has been expanded to natural language processing such as intent detection~\cite{naacl/Retriever, protoda-fewshot}.

Few-shot classification is an important and challenging task. Several methods have been proposed to tackle this problem. In particular, several metric-based methods \cite{vinyals2016matching, snell2017prototypical, few-shot-text, EMNLP19, iclr/BaoWCB20} have been proposed, which first learn an embedding space and then utilize a metric to classify instances of new categories according to proximities with the labeled examples. In addition to metric-based methods, some optimization-based approaches ~\cite{iclr/RaviL17, finn2017model,DBLP:conf/nips/YoonKDKBA18} have also been explored for few-shot classification.

\paragraph{Contrastive learning}
Contrastive learning applied to self-supervised representation learning has seen a resurgence of interest in recent years, leading to state-of-the-art performance in unsupervised training of deep image models \cite{DBLP:conf/icml/ChenK0H20}. \citet{DBLP:conf/nips/KhoslaTWSTIMLK20} extend the self-supervised batch contrastive approach to the fully-supervised setting, allowing us to effectively leverage label information. Recently, \citet{DBLP:journals/corr/abs-2011-01403} propose a novel objective function that contains a supervised contrastive learning term for fine-tuning pre-trained language models, which improves the model generalization ability significantly.

\paragraph{Joint intent classification and slot filling} Due to the close relationship between IC and SF, \citet{interspeech/LiuL16, ijcai/ZhangW16a, naacl/GooGHHCHC18, emnlp/QinCLWL19, co-interactive} propose joint models to consider the correlation between these two tasks. These models can be classified into two categories. One type of approaches \cite{interspeech/LiuL16, ijcai/ZhangW16a} adopt a multi-task framework to solve these two tasks simultaneously. Although these models outperform the single-task model, they just model the relationship implicitly by sharing the encoder parameters. The other type of approaches \cite{naacl/GooGHHCHC18, emnlp/QinCLWL19} explicitly adopt the intent information to guide the slot filling task. \citet{co-interactive} further propose a co-interactive transformer which considers the cross-impact between these two tasks. These explicit-joint learning models have achieved very remarkable performance, but they mainly focus on the traditional supervised learning setting.

\section{Problem Definition}
A labeled utterance with $ T $ words (tokens) can be represented as $ (x, t, y)$, where $ x = (w_1, w_2, ... , w_T) $ is an utterance with $T$ words, $ t = (t_1, t_2, ... , t_T) $ is composed of slot labels of each word in $x$, $ y $ is the intent label of $x$. In this paper, few-shot classification is conducted via episode learning strategy. In the training period, we partition the training set into multiple episodes. Each episode consists of a \textit{support set} $\mathcal{S}$ and a \textit{query set} $\mathcal{Q}$. In particular, we randomly select $N$ classes from the training classes, and obtain a class set $\mathcal{C}$ in each episode. Then the support set is formed by randomly selecting $k_c$ labeled samples (utterances) from each of the $N$ classes, i.e., $\mathcal{S} = \bigcup_{c \in\mathcal{C}}\mathcal{S}_c$, where $\mathcal{S}_c = \{ (x^i, t^i, y_c)|i\in(1,...,k_c)\}$. And a fraction of the remainder of these $N$ classes' samples ($k_q$ examples per class) serve as the query set, i.e., $ \mathcal{Q} = \bigcup_{c \in\mathcal{C}}\mathcal{Q}_c $, where $ \mathcal{Q}_c = \{ (x^j, t^j, y_c)|j\in(1,...,k_q)\}$. In the test period, we also partition the test set into multiple episodes. Each episode contains a support set $\mathcal{S} = \bigcup_{c \in\mathcal{C}}\mathcal{S}_c$, where $\mathcal{S}_c = \{ (x^i, t^i, y_c)|i\in(1,...,k_c)\}$, and a query set  $ \mathcal{Q} = \bigcup_{c \in\mathcal{C}}\mathcal{Q}_c $, where  $ \mathcal{Q}_c = \{ x^j|j\in(1,...,k_q)\}$. There is no overlap between the training classes and test classes. Table \ref{Tab.1} summarizes the symbol explanation in details.

\begin{table}
    \small
	\centering
	\linespread{2}
	\begin{tabular}{|c|l|}
		\hline
		\multicolumn{1}{|c|}{Symbol} & \multicolumn{1}{c|}{Explanation}  \\
		\hline
		$ \mathcal{C}$       			&set of intent classes in each episode      \\
		$ \mathcal{S} $					&support set of an episode				    \\
		$ \mathcal{Q} $					&query set of an episode                    \\
		$ \mathcal{S}_c $               &set of support data in the $c$-th class	\\
		$\mathcal{Q}_c $                &set of  query data in the $c$-th class	    \\
		$ x $                           &an utterance with $ T $ words, $x= (w_1, ... , w_T)$             \\
		$ t $ 		                    &slot labels of each word in $x$, $t=(t_1,  ... , t_T)$          \\
		$ y$ 							&intent label of utterance $x$ 	        \\
		$ k_c$ 							&number of supports in $ \mathcal{S}_c $        \\
		$ k_q$ 							&number of queries in $\mathcal{Q}_c $ \\
		$ \textit{\textbf{H}}$ 			&pre-processed utterance embedding          \\
		$ \textit{\textbf{E}}_I$ 		&intent label embedding \\
		$ \textit{\textbf{E}}_S$ 		&slot label embedding \\
		$ \textit{\textbf{H}}_I$   		&slot-attention-based intent representation           \\
		$ \textit{\textbf{H}}_S$        &intent-attention-based slot representation           \\
		$ \textit{\textbf{c}} $ 		&sentence embedding of utterance $ x $      \\
    \hline
	\end{tabular}
	\caption{Symbol explanation.}
	\label{Tab.1}
\end{table}

\section{Approach}
\subsection{Pre-processing}
Given an utterance $ x = (w_1, w_2, ... , w_T) $ with $T$ words (tokens), each word in the utterance can obtain its word embedding by BERT \cite{naacl/DevlinCLT19}. And each word can be further encoded using a recurrent neural network such as bidirectional LSTM, i.e.,
\begin{equation}
\begin{split}
& \overrightarrow{\textit{\textbf{h}}}_t = {\rm LSTM}_{fw}(\textit{{w}}_t, \overrightarrow{\textit{\textbf{h}}}_{t-1}), \\
& \overleftarrow{\textit{\textbf{h}}}_t = {\rm LSTM}_{bw}(\textit{{w}}_t, \overleftarrow{\textit{\textbf{h}}}_{t+1}),
\end{split}
\end{equation}
where LSTM$_{fw}$ and LSTM$_{bw}$ denote the forward and backward LSTM respectively, and $ \overrightarrow{\textit{\textbf{h}}}_t \in\mathbb{R}^{d_h}$ and $ \overleftarrow{\textit{\textbf{h}}}_t \in\mathbb{R}^{d_h} $ are the hidden states of the $t$-th word learned from LSTM$_{fw}$ and LSTM$_{bw}$ respectively. The entire hidden state of the $t$-th word is represented by concatenating $ \overrightarrow{\textit{\textbf{h}}}_t $ and $ \overleftarrow{\textit{\textbf{h}}}_t $, i.e., $ \textit{\textbf{h}}_t = [\overrightarrow{\textit{\textbf{h}}}_t , \overleftarrow{\textit{\textbf{h}}}_t] $, and the hidden state matrix of the utterance is $ \textit{\textbf{H}} = [\textit{\textbf{h}}_1, \textit{\textbf{h}}_2, ... , \textit{\textbf{h}}_T]^\top \in\mathbb{R}^{ T\times 2d_{h}}$. To express concisely, we use $ d = 2d_h $ to represent the dimension of hidden state and obtain $  \textit{\textbf{H}} \in\mathbb{R}^{T \times d}$.

\subsection{Extracting Intent and Slot Representations via Bidirectional Interaction}
To explicitly establish the interaction between intent classification and slot filling, for each utterance, we first use the attention mechanism over slot and intent label descriptions to get the initial intent and slot representations  \cite{emnlp/CuiZ19,co-interactive}. Then, these initial representations are concatenated with the utterance embedding matrix to produce the final slot-attention-based intent representation and intent-attention-based slot representation.

In particular, we first use the embeddings of intent labels' descriptions to produce intent embedding matrix $\textbf{\textit{E}}_I\in\mathbb{R}^{|\mathcal{C}_{intent}|\times d}$, and use the embeddings of slot labels' descriptions to produce slot embedding matrix $\textbf{\textit{E}}_S\in\mathbb{R}^{|\mathcal{C}_{slot}|\times d}$, where $ |\mathcal{C}_{intent}| $ is the number of intents in the episode, $ |\mathcal{C}_{slot}| $ is the number of slots in the episode, $ d $ is the dimension of hidden state. $\textbf{\textit{E}}_I$ and $ \textbf{\textit{E}}_S$ are initialized by pre-processing intent and slot labels' descriptions, and they are learnable and can be updated during training. Then we calculate slot-attention-based intent representation and intent-attention-based slot representation as follows.

\paragraph{Slot-attention-based Intent Representation}
\begin{equation}
	\textbf{\textit{H}}_I =  softmax(\textbf{\textit{H}} (\textbf{\textit{E}}_S)^T)\textbf{\textit{E}}_S \ || \ \textbf{\textit{H}}.
\end{equation}

\paragraph{Intent-attention-based Slot Representation}
\begin{equation}
\textbf{\textit{H}}_S =  softmax(\textbf{\textit{H}} (\textbf{\textit{E}}_I)^T)\textbf{\textit{E}}_I \ || \ \textbf{\textit{H}}.
\end{equation}
Here $ \textbf{\textit{H}}_I = (\textit{\textbf{h}}_{1}^{I}, \textit{\textbf{h}}_{2}^{I}, ..., \textit{\textbf{h}}_{T}^{I}) \in\mathbb{R}^{T \times 2d}$, $ \textbf{\textit{H}}_S = (\textit{\textbf{h}}_{1}^{S}, \textit{\textbf{h}}_{2}^{S}, ..., \textit{\textbf{h}}_{T}^{S}) \in\mathbb{R}^{T \times 2d} $, and they carry the corresponding intent and slot information respectively.

\subsection{Explicit Joint Learning with Prototypical Networks}
Inspired by \cite{handful-icsf}, we also extend the prototypical networks to perform joint intent classification and slot filling. Different from \cite{handful-icsf}, when calculating the prototype of slot label, instead of only considering the words in the front, we use the window strategy to take the contextual words into account simultaneously, which seems more reasonable.

In general, for each intent class or slot class, its corresponding prototype is the mean vector of the sample embeddings in that class. Given a support set, $ \mathcal{S}_c = \{ (x^i, t^i, y_c)|i\in(1,...,k_c)\}$ is the set of support data with intent class $c$, where $ x^i=(w_1^i, w_2^i, ..., w_T^i) $ is the $i$-th utterance, and $ t^i=(t_1^i, t_2^i, ..., t_T^i) $ is the corresponding slot labels. $ \mathcal{S}_o = \{(x^i, t^i, y^i)|t_j^i = o\} $ is the set of support data with slot label $ o $. The prototype $ \textit{\textbf{p}}_c $ of intent label $ c $ and the prototype $ \textit{\textbf{p}}_o $ of slot label $ o $ can be computed as follows:
\begin{equation}
\textit{\textbf{p}}_c=\frac{1}{\left|\mathcal{S}_c\right|}\sum_{x^i\in \mathcal{S}_c}\textit{\textbf{c}}^i,
\end{equation}
\begin{equation}
\textit{\textbf{p}}_o=\frac{1}{\left|\mathcal{S}_o\right|}\sum_{x^i \in \mathcal{S}_o}{\frac{1}{2l+1}\sum_{k=j-l}^{j+l} (\textit{\textbf{h}}_k^S)^i},
\end{equation}
where $ \textit{\textbf{c}}^i = mean(\textit{\textbf{H}}_I)\in\mathbb{R}^{2d}$ is the embedding of the $i$-th utterance $x^i$. ${\frac{1}{2l+1}\sum_{k=j-l}^{j+l} (\textit{\textbf{h}}_k^S)^i}$ is the embedding of $j$-th word with slot label $o$, which considers the contextual words simultaneously with the window size $2l+1$.

Given a query data $ (x^*, t^*, y^*)  \in\mathcal{Q} $, we compute the conditional probability $ p\left(y=\left.c\right|x^*,\ \mathcal{S}\right) $ to predict its intent based on negative squared Euclidean distance.
\begin{equation}
p\left(y\!=\!\left.c\right|x^*,\mathcal{S}\right) = \frac{exp(-{||\textit{\textbf{c}}^*-\textit{\textbf{p}}_c||}_2^2)}{\sum_{c'}exp(-{||\textit{\textbf{c}}^*-\textit{\textbf{p}}_{c'}||}_2^2)}.
\end{equation}
Here $\textit{\textbf{c}}^*$ is the embedding of $x^*$. Similarly, we can compute the conditional probability $ p\left(t_j=\left.o\right|x^*, \ \mathcal{S}\right) $ to predict the slot. 

Finally, we perform the cross-entropy loss on all query instances to construct the IC and SF prototypical loss functions.

\begin{equation}
\mathcal{L}_{IC_{pn}} = \frac{1}{\left|\mathcal{Q}\right|}\sum_{x^*\in \mathcal{Q}} -\log{p\left(y=\left.y^\ast\right|x^\ast,\mathcal{S}\right)}.
\label{Eq.7}
\end{equation}
\begin{equation}
\mathcal{L}_{SF_{pn}} = \frac{1}{\left|\mathcal{Q}\right|} \sum_{x^*\in \mathcal{Q}} \sum_{t_j^\ast\in t^\ast}{ - log p \left( t_j=\left.t_j^\ast\right|x^\ast,\mathcal{S}\right)}.
\label{Eq.8}
\end{equation}

\subsection{Integrating with Supervised Contrastive Learning}
Supervised contrastive learning has achieved great success in computer vision, which aims to maximize similarities between instances from the same class and minimize similarities between instances from different classes. Here we integrate with supervised contrastive learning to generate better intent representations and slot representations. 

We first construct contrastive samples for each query instance using support set. For a query instance $ x $, we can take the support instances which have the same label with $x$ as the positive samples, and the negative samples are those with different labels. Then for an episode, the SCL loss of IC can be written as:
\begin{equation}
	\begin{split}
		\mathcal{L}_{IC_{scl}}=\frac{1}{\left|\mathcal{Q}\right|}&\sum_{x^i\in{\mathcal{Q}}}-\frac{1}{N_{y^i}}\sum_{x^j\in{\mathcal{S}}} \mathbf{1}_{y^i=y^j} \\
			&log\ \frac{exp({\textbf{z}}^i\cdot{\textbf{z}}^j/\tau)\ }{\sum_{x^k\in{\mathcal{S}}}{exp({\textbf{z}}^i\cdot{\textbf{z}}^k/\tau)}},
	\end{split}
	\label{Eq.9}
\end{equation}
where ${\textbf{z}}^i\cdot{\textbf{z}}^j$ means the inner product of the two vectors. $ (x^i,\ t^i,\ y^i) $ is a query instance in query set $ \mathcal{Q} $. $ N_{y^i} $ is the total number of utterances in support set which have the same intent label $ y^i $. ${\textbf{z}}^i = mean(\textit{\textbf{H}})$ is the pre-processed embedding of $x^i$. $ \tau > 0 $ is an adjustable scalar parameter which can control the separation degree of classes.

To analyze Eq.~(\ref{Eq.9}), we can do some simple formula manipulation as below. 

\begin{small}
\begin{equation*}
	\begin{split}
		& \mathcal{L}_{IC_{scl}}=\frac{1}{\left|\mathcal{Q}\right|}\sum_{x^i\in{\mathcal{Q}}}-\frac{1}{N_{y^i}}\mathcal{L}_{scl}, \\
		& \mathcal{L}_{scl}=\sum_{x^j\in{\mathcal{S}}} \mathbf{1}_{y^i=y^j} log\ \frac{exp({\textbf{z}}^i\cdot{\textbf{z}}^j/\tau)\ }{\sum_{x^k\in{\mathcal{S}}}{exp({\textbf{z}}^i\cdot{\textbf{z}}^k/\tau)}} \\
		& \!=\!\underbrace{\sum_{x^j\in{\mathcal{S}}} \! \mathbf{1}_{y^i=y^j}\!(\frac{{\textbf{z}}^i\cdot{\textbf{z}}^j}{\tau})}_{positive}\! - \!
		\!\sum_{x^j\in{\mathcal{S}}}\!\mathbf{1}_{y^i=y^j} log \! \underbrace{\sum_{x^k\in{\mathcal{S}}}{exp(\frac{{\textbf{z}}^i\cdot{\textbf{z}}^k}{\tau})}}_{positive+negative}\!.
	\end{split}		
\end{equation*}
\end{small}

According to the above formula, if we want to minimize $ \mathcal{L}_{IC_{scl}} $, we must maximize $ \mathcal{L}_{scl}$, where we need to maximize the \textit{positive} term and minimize the \textit{positive+negative} term, so the \textit{negative} term will be decreased. Intuitively, the supervised contrastive learning term can push samples from the same class close and samples from different classes further apart.

In a similar manner, the SCL loss of SF for an episode can be written as:
\begin{equation}
	\begin{split}
		\mathcal{L}_{SF_{scl}}&=\frac{1}{\left|\mathcal{Q}_s\right|}\sum_{w_i\in{\mathcal{Q}_s}}-\frac{1}{N_{t_i}}\sum_{ w_j \in{\mathcal{S}_s}} \\
		&{\mathbf{1}_{t_i=t_j}log\ \frac{exp(\textit{\textbf{h}}_i\cdot\textit{\textbf{h}}_j/\tau)\ }{\sum_{w_k\in{\mathcal{S}_s}}{exp(\textit{\textbf{h}}_i\cdot\textit{\textbf{h}}_k/\tau)}}},
	\end{split}
	\label{Eq.11}
\end{equation}
where $\textit{\textbf{h}}_i$ and $\textit{\textbf{h}}_j$ are the embedding representations of $w_i$ and $w_j$. $\textit{\textbf{h}}_i\cdot\textit{\textbf{h}}_j$ means the inner product of $\textit{\textbf{h}}^i$ and $\textit{\textbf{h}}^j$. $ \mathcal{Q}_s $ represents the set of words in query set, and $ \mathcal{S}_s $ represents the set of words in support set. $ N_{t_i} $ is the total number of words in support set which have the same slot label $ t_i $.  Here the same word in different utterances are considered repeatedly, and the words with slot label "Other" are ignored. Note that different from the symbol $t^i$ which represents the slot labels of each word in an utterance $x^i$, $t_i$ represents the slot label of $w_i$.

Combining Eq.~(\ref{Eq.7}), (\ref{Eq.8}), (\ref{Eq.9}) and (\ref{Eq.11}), the overall loss function of the proposed framework is:
\begin{equation}
	\mathcal{L} = \mathcal{L}_{IC_{pn}} + \lambda \mathcal{L}_{SF_{pn}} + \gamma {\mathcal{L}_{IC_{scl}}} + \delta{\mathcal{L}}_{SF_{scl}},
\end{equation}
where $ \lambda, \gamma$ and $ \delta $ are trade-off hyperparameters.

\section{Experiments}
\subsection{Episode Construction}
In this section, we outline the method of sampling episodes used in \cite{iclr/TriantafillouZD20} and \cite{handful-icsf}, which allows that the "way" $N$ and the "shot" $k_c$ are variable in each episode, and can cater the unbalanced datasets and very limited labeled instances in real application scenarios. Given a data split which contains $ |C_{split}| $ intent classes, there are two steps to construct an episode.

\paragraph{Step 1:} Sampling the class set for each episode.

\noindent\textbf{(i)} We sample the class number $ N $ uniformly from the range [3, $ |C_{split}| $].

\noindent\textbf{(ii)} We sample $ N $ intent classes from the data split at random.

\paragraph{Step 2:} Sampling the samples for each episode.

\noindent\textbf{(i)} Computing the query set size of each class by:
\begin{equation*}
k_{q} =\min\{10, (\underset{c\in \mathcal{C}}{\min} \left\lfloor0.5\ast\left|U(c)\right|\right\rfloor)\},
\end{equation*}
where $ \mathcal{C} $ is the set of selected classes, and $ U(c) $ denotes the set of utterances belonging to class $ c $.

\noindent\textbf{(ii)} Computing the total support set size $|\mathcal{S}|$ by:
\begin{equation*}
\min
\left\{
	U_{max}, \sum_{c\in{\mathcal{C}}}\lceil\beta\min \{20, \left|U(c)\right| - k_q\}\rceil
\right\},
\end{equation*}
where $ \beta $ is a scalar sampled uniformly from interval $ (0,1] $, and $ U_{max}$ is the maximum support set size.

\noindent\textbf{(iii)} Computing the number of shots $ k_c $ of each class by:
\begin{equation*}
k_c = \min \left\{ \left\lfloor R_c \ast (|\mathcal{S}| - |\mathcal{C}|)\right\rfloor + 1, |U(c)| - k_q
\right\},
\end{equation*}
where the parameter $R_c$ is computed by:
\begin{equation*}
R_c = \frac{exp(\alpha_c)|U(c)|}{\sum_{{c'}\in \mathcal{C}}exp(\alpha_{c'}) |U(c')|},
\end{equation*}
where $ \alpha_c $ is sampled uniformly from the interval $ [log(0.5), log(2)) $.

\begin{table}[b]
    \scriptsize
	\centering
	\setlength\tabcolsep{8pt}
	\begin{tabular}{|l|cc|cc|cc|}
		\hline
		\multicolumn{1}{|c|}{\multirow{2}{*}{Split}} & \multicolumn{2}{c|}{ATIS} & \multicolumn{2}{c|}{SNIPS}  & \multicolumn{2}{c|}{TOP}  \\
		\multicolumn{1}{|c|}{}& \multicolumn{1}{l}{\#Utt} & \multicolumn{1}{l|}{\#In} & \multicolumn{1}{l}{\#Utt} & \multicolumn{1}{l|}{\#In} & \multicolumn{1}{l}{\#Utt} & \multicolumn{1}{l|}{\#In} \\ \hline
		Train               & 4,373  & 5  & 8,230  & 4  & 20,345 & 7 \\
		Dev                 & 669    & 6  & -      & -  & 4,333  & 5 \\
		Test                & 829    & 7  & 6,254  & 3  & 4,426  & 6 \\ \hline
		Total               & 5,871  & 18 & 14,484 & 7  & 29,104 & 18\\ \hline
	\end{tabular}
	\caption{Detailed statistics on utterance (Utt) and intent (In) counts for ATIS, SNIPS and TOP.}
	\label{tab:stat}
\end{table}

\begin{table*}[t]
	\centering
	\resizebox{\textwidth}{33mm}{
	\begin{tabular}{|ll|l|l|l|l|l|l|}
				\hline
				\multicolumn{1}{|c}{\multirow{2}{*}{Embed.}} & \multicolumn{1}{c|}{\multirow{2}{*}{Algorithm}} & \multicolumn{6}{c|}{IC Accuracy (mean +/- std)}                                                                                                                                                           \\ \cline{3-8}
				\multicolumn{1}{|c}{} & \multicolumn{1}{c|}{}  & \multicolumn{1}{c|}{SNIPS} & \multicolumn{1}{c|}{SNIPS (joint)} & \multicolumn{1}{c|}{ATIS} & \multicolumn{1}{c|}{ATIS (joint)} & \multicolumn{1}{c|}{TOP} & \multicolumn{1}{c|}{TOP (joint)} \\ \hline
				GloVe     & Fine-tune  & 69.52 +/- 2.88  & 70.25 +/- 1.85  & 49.50 +/- 0.65  & 58.26 +/- 1.12 & 37.58 +/- 0.54  & 40.93 +/- 2.77  \\
				GloVe     & foMAML     & 61.08 +/- 1.50  & 59.67 +/- 2.12  & 54.66 +/- 1.82  & 45.20 +/- 1.47 & 33.75 +/- 1.30  & 31.48 +/- 0.50  \\
				GloVe     & Proto      & 68.19 +/- 1.76  & 68.77 +/- 1.60  & 65.46 +/- 0.81  & 63.91 +/- 1.27 & 43.20 +/- 0.85  & 38.65 +/- 1.35  \\ \hline
				ELMo      & Fine-tune  & 85.53 +/- 0.35  & \textbf{87.64 +/- 0.73}  & 49.25 +/- 0.74  & 58.69 +/- 1.56   & 45.49 +/- 0.61  & 47.63 +/- 2.75   \\
				ELMo      & foMAML     & 78.90 +/- 0.77  & 78.86 +/- 1.31  & 53.90 +/- 0.96  & 52.47 +/- 2.86 & 38.67 +/- 1.02  & 36.49 +/- 0.99  \\
				ELMo      & Proto      & 83.54 +/- 0.40  & 85.75 +/- 1.57  & 65.95 +/- 2.29  & 65.19 +/- 1.29 & 50.57 +/- 2.81  & 50.64 +/- 2.72  \\ \hline
				BERT      & Fine-tune  & 76.04 +/- 8.84  & 77.53 +/- 5.69  & 43.76 +/- 4.61  & 50.73 +/- 3.86 & 39.21 +/- 3.09  & 40.86 +/- 3.75  \\
				BERT      & foMAML     & 67.36 +/- 1.03  & 68.37 +/- 0.48  & 50.27 +/- 0.69  & 48.80 +/- 2.82 & 38.50 +/- 0.43  & 36.20 +/- 1.21  \\
				BERT      & Proto      & 81.39 +/- 1.85  & 81.44 +/- 2.91  & 58.84 +/- 1.33  & 58.82 +/- 1.55 & 52.76 +/- 2.26  & 52.64 +/- 2.58  \\ \hline
				\multicolumn{2}{|l|}{Retriever}     & \multicolumn{2}{c|}{68.81 +/- 0.32}   & \multicolumn{2}{c|}{49.22 +/- 0.79} & \multicolumn{2}{c|}{50.67 +/- 0.44}
	            \\ \hline
				\multicolumn{2}{|l|}{our framework (o, o)}          & \multicolumn{2}{c|}{84.61 +/- 0.78}  & \multicolumn{2}{c|}{76.09 +/- 3.75} & \multicolumn{2}{c|}{59.63 +/- 1.48} \\
				\multicolumn{2}{|l|}{our framework (w, o)}  & \multicolumn{2}{c|}{\textbf{85.81 +/- 0.45}}  & \multicolumn{2}{c|}{\textbf{80.37 +/- 0.58}} & \multicolumn{2}{c|}{\textbf{62.81 +/- 0.96}}        \\
				\multicolumn{2}{|l|}{our framework (w, w)}  & \multicolumn{2}{c|}{85.15 +/- 0.67} & \multicolumn{2}{c|}{\textbf{80.44 +/- 0.62}} & \multicolumn{2}{c|}{\textbf{62.85 +/- 0.33}} \\ \hline
	\end{tabular}}
	\caption{Average IC accuracy on 100 test episodes when $ U_{max} = 20 $.}
	\label{tab:0acc}
\end{table*}

\begin{table*}[t]
	\centering
	\resizebox{\textwidth}{33mm}{
	\begin{tabular}{|ll|l|l|l|l|l|l|}
				\hline
				\multicolumn{1}{|c}{\multirow{2}{*}{Embed.}} & \multicolumn{1}{c|}{\multirow{2}{*}{Algorithm}} & \multicolumn{6}{c|}{IC Accuracy (mean +/- std)}                                                                                                                                                           \\ \cline{3-8}
				\multicolumn{1}{|c}{} & \multicolumn{1}{c|}{} & \multicolumn{1}{c|}{SNIPS} & \multicolumn{1}{c|}{SNIPS (joint)} & \multicolumn{1}{c|}{ATIS} & \multicolumn{1}{c|}{ATIS (joint)} & \multicolumn{1}{c|}{TOP} & \multicolumn{1}{c|}{TOP (joint)} \\ \hline
				GloVe    & Fine-tune     & 72.24 +/- 2.58   & 73.00 +/- 1.84  & 49.91 +/- 1.90 & 56.07 +/- 2.94 & 39.66 +/- 1.34 & 41.10 +/- 0.65  \\
				GloVe    & foMAML        & 66.75 +/- 1.28   & 67.34 +/- 2.62  & 54.92 +/- 0.87 & 58.46 +/- 1.91 & 33.62 +/- 1.53 & 35.68 +/- 0.62  \\
				GloVe    & Proto         & 70.45 +/- 0.49   & 72.66 +/- 1.96  & 70.25 +/- 0.39 & 69.58 +/- 0.41 & 48.84 +/- 1.59 & 46.85 +/- 0.86  \\ \hline
				ELMo     & Fine-tune     & 87.69 +/- 1.05   & \textbf{88.90 +/- 0.18}   & 49.42 +/- 0.79    & 56.99 +/- 2.12  & 47.44 +/- 1.61  & 48.87 +/- 0.54   \\
				ELMo     & foMAML        & 80.80 +/- 0.47   & 81.62 +/- 1.07            & 59.10 +/- 2.52    & 56.16 +/- 1.34  & 41.80 +/- 1.49  & 36.24 +/- 0.79   \\
				ELMo     & Proto         & 86.76 +/- 1.62   & \textbf{87.74 +/- 1.08}   & 70.10 +/- 1.26    & 71.89 +/- 1.45  & 58.60 +/- 1.91  & 56.87 +/- 0.39   \\ \hline
				BERT     & Fine-tune     & 76.66 +/- 8.68   & 79.53 +/- 4.25            & 44.08 +/- 6.05    & 49.71 +/- 3.84  & 40.05 +/- 2.35  & 40.46 +/- 1.74   \\
				BERT     & foMAML        & 70.43 +/- 1.56   & 72.79 +/- 1.11            & 51.36 +/- 3.74    & 50.25 +/- 0.88  & 36.15 +/- 2.17  & 35.24 +/- 0.35   \\
				BERT     & Proto         & 83.51 +/- 0.88   & 86.29 +/- 1.09            & 66.89 +/- 2.31    & 65.70 +/- 2.31  & 61.30 +/- 0.32  & 62.51 +/- 1.79   \\ \hline
				\multicolumn{2}{|l|}{Retriever}   & \multicolumn{2}{c|}{71.98 +/- 0.42}  & \multicolumn{2}{c|}{54.79 +/- 0.27}   & \multicolumn{2}{c|}{51.78 +/- 0.61} \\ \hline
				\multicolumn{2}{|l|}{our framework (o, o)} & \multicolumn{2}{c|}{86.35 +/- 1.32}  & \multicolumn{2}{c|}{84.92 +/- 1.75}   & \multicolumn{2}{c|}{67.98 +/- 1.21} \\
				\multicolumn{2}{|l|}{our framework (w, o)} & \multicolumn{2}{c|}{86.46 +/- 0.89}  & \multicolumn{2}{c|}{\textbf{86.85 +/- 0.59}} 
& \multicolumn{2}{c|}{\textbf{68.74 +/- 0.61}}  \\
				\multicolumn{2}{|l|}{our framework (w, w)} & \multicolumn{2}{c|}{86.79 +/- 0.37} & \multicolumn{2}{c|}{\textbf{86.29 +/- 0.42}} & \multicolumn{2}{c|}{\textbf{68.51 +/- 0.77}} \\ \hline
	\end{tabular}}
	\caption{Average IC accuracy on 100 test episodes when $ U_{max} = 100 $.}
	\label{tab:1acc}
\end{table*}

\begin{table*}[t]
	\centering
	\resizebox{\textwidth}{33mm}{
	\begin{tabular}{|ll|l|l|l|l|l|l|}
				\hline
				\multicolumn{1}{|c}{\multirow{2}{*}{Embed.}} & \multicolumn{1}{c|}{\multirow{2}{*}{Algorithm}} & \multicolumn{6}{c|}{SF F1 Score (mean +/- std)}                                                                                                                                                           \\ \cline{3-8}
				\multicolumn{1}{|c}{}  & \multicolumn{1}{c|}{} & \multicolumn{1}{c|}{SNIPS} & \multicolumn{1}{c|}{SNIPS (joint)} & \multicolumn{1}{c|}{ATIS} & \multicolumn{1}{c|}{ATIS (joint)} & \multicolumn{1}{c|}{TOP} & \multicolumn{1}{c|}{TOP (joint)} \\ \hline
				GloVe    & Fine-tune      & 6.72 +/- 1.24  & 6.68 +/- 0.40  & 2.57 +/- 1.21  & 13.22 +/- 1.07 & 0.90 +/- 0.51   & 0.76 +/- 0.21  \\
				GloVe    & foMAML         & 14.07 +/- 1.01 & 12.91 +/- 0.43 & 18.44 +/- 0.91 & 16.91 +/- 0.32 & 5.34 +/- 0.43   & 9.22 +/- 1.03  \\
				GloVe    & Proto          & 29.63 +/- 0.75 & 27.75 +/- 2.52 & 31.19 +/- 1.15 & 38.45 +/- 0.97 & 10.65 +/- 0.83  & 18.55 +/- 0.35 \\ \hline
				ELMo     & Fine-tune      & 22.02 +/- 1.13 & 16.00 +/- 2.07 & 7.47 +/- 2.60  & 7.19 +/- 1.71  & 1.26 +/- 0.46   & 1.17 +/- 0.32  \\
				ELMo     & foMAML         & 33.81 +/- 0.33 & 32.82 +/- 0.84 & 27.58 +/- 1.25 & 24.45 +/- 1.20 & 22.35 +/- 1.23  & 15.53 +/- 0.64 \\
				ELMo     & Proto          & \textbf{59.88 +/- 0.53}  & \textbf{59.73 +/- 1.72}  & 33.97 +/- 0.38  & 40.90 +/- 2.21  & 20.12 +/- 0.25 & 28.97 +/- 0.82  \\ \hline
				BERT     & Fine-tune      & 12.47 +/- 0.31           & 8.75 +/- 0.28            & 9.24 +/- 1.67   & 15.93 +/- 3.10  & 3.15 +/- 0.28  & 1.08 +/- 0.30   \\
				BERT     & foMAML         & 12.72 +/- 0.12           & 13.28 +/- 0.53           & 18.91 +/- 1.01  & 16.05 +/- 0.32  & 5.93 +/- 0.43  & 8.23 +/- 0.81   \\
				BERT     & Proto          & 42.09 +/- 1.11           & 43.77 +/- 0.54           & 37.61 +/- 0.82  & 39.27 +/- 1.84  & 20.81 +/- 0.40 & 28.24 +/- 0.53  \\ \hline
				\multicolumn{2}{|l|}{Retriever}  & \multicolumn{2}{c|}{48.30 +/- 0.05}  & \multicolumn{2}{c|}{\textbf{64.14 +/- 0.99}}   & \multicolumn{2}{c|}{34.77 +/- 0.34}  \\ \hline
				\multicolumn{2}{|l|}{our framework (o, o)}       & \multicolumn{2}{c|}{50.03 +/- 0.59}  & \multicolumn{2}{c|}{61.79 +/- 3.06}   & \multicolumn{2}{c|}{38.41 +/- 1.02}  \\
				\multicolumn{2}{|l|}{our framework (w, o)}  & \multicolumn{2}{c|}{50.77 +/- 0.92}  & \multicolumn{2}{c|}{62.73 +/- 0.53 }  &
\multicolumn{2}{c|}{\textbf{38.82 +/- 0.87}}      \\
				\multicolumn{2}{|l|}{our framework (w, w)} & \multicolumn{2}{c|}{52.82 +/- 0.70} & \multicolumn{2}{c|}{\textbf{63.65 +/- 0.55}} &
\multicolumn{2}{c|}{\textbf{39.92 +/- 0.42}}      \\ \hline
			\end{tabular}	
	}
	\caption{Average SF F1 score on 100 test episodes when $ U_{max} = 20 $.}
	\label{tab:0f1}
\end{table*}

\begin{table*}[t]
	\centering
	\resizebox{\textwidth}{33mm}{
	\begin{tabular}{|ll|l|l|l|l|l|l|}
				\hline
				\multicolumn{1}{|c}{\multirow{2}{*}{Embed.}} & \multicolumn{1}{c|}{\multirow{2}{*}{Algorithm}} & \multicolumn{6}{c|}{SF F1 Score (mean +/- std)}                                                                                                                                       \\ \cline{3-8}
				\multicolumn{1}{|c}{}  & \multicolumn{1}{c|}{}  & \multicolumn{1}{c|}{SNIPS} & \multicolumn{1}{c|}{SNIPS (joint)} & \multicolumn{1}{c|}{ATIS} & \multicolumn{1}{c|}{ATIS (joint)} & \multicolumn{1}{c|}{TOP} & \multicolumn{1}{c|}{TOP (joint)} \\ \hline
				GloVe     & Fine-tune  & 7.06 +/- 1.87  & 7.76 +/- 0.91   & 2.72 +/- 1.65  & 17.20 +/- 3.03  & 1.26 +/- 0.44  & 0.67 +/- 0.33  \\
				GloVe     & foMAML     & 16.77 +/- 0.67 & 16.53 +/- 0.32  & 17.80 +/- 0.42 & 23.33 +/- 2.89  & 4.11 +/- 0.81  & 9.89 +/- 1.13  \\
				GloVe     & Proto      & 31.57 +/- 1.28 & 31.17 +/- 1.31  & 31.32 +/- 2.79 & 41.07 +/- 1.14  & 9.99 +/- 1.08  & 18.93 +/- 0.77 \\ \hline
				ELMo      & Fine-tune  & 22.37 +/- 0.91 & 17.09 +/- 2.57  & 8.93 +/- 2.86  & 11.09 +/- 2.00  & 2.04 +/- 0.41  & 1.03 +/- 0.24  \\
				ELMo      & foMAML     & 36.10 +/- 1.49 & 37.33 +/- 0.24  & 26.91 +/- 2.64 & 26.37 +/- 0.15  & 18.32 +/- 0.52 & 16.55 +/- 0.79 \\
				ELMo      & Proto      & \textbf{62.71 +/- 0.40}  & \textbf{62.14 +/- 0.75}  & 35.20 +/- 2.46 & 41.28 +/- 2.73  & 18.44 +/- 2.41 & 28.33 +/- 1.33  \\ \hline
				BERT      & Fine-tune  & 14.71 +/- 0.43           & 10.50 +/- 0.90           & 11.53 +/- 1.46 & 20.41 +/- 1.85  & 4.98 +/- 0.66  & 1.48 +/- 0.85   \\
				BERT      & foMAML     & 14.99 +/- 1.29           & 15.83 +/- 0.94           & 17.68 +/- 2.42 & 17.11 +/- 1.31  & 3.37 +/- 0.36  & 10.58 +/- 0.45  \\
				BERT      & Proto      & 46.50 +/- 0.75           & 48.77 +/- 0.71           & 40.63 +/- 3.37 & 43.10 +/- 1.76  & 20.58 +/- 2.27 & 28.92 +/- 1.09  \\ \hline
				\multicolumn{2}{|l|}{Retriever} & \multicolumn{2}{c|}{49.39 +/- 0.78} & \multicolumn{2}{c|}{\textbf{68.13 +/- 3.06}} & \multicolumn{2}{c|}{37.12 +/- 0.84}  \\ \hline
				\multicolumn{2}{|l|}{our framework (o, o)}      & \multicolumn{2}{c|}{54.29 +/- 0.99} & \multicolumn{2}{c|}{59.13 +/- 1.69}  & \multicolumn{2}{c|}{\textbf{38.74 +/- 1.53}} \\
				\multicolumn{2}{|l|}{our framework (w, o)}  & \multicolumn{2}{c|}{54.52 +/- 0.31} & \multicolumn{2}{c|}{62.01 +/- 0.50} &
\multicolumn{2}{c|}{38.40 +/- 0.21}    \\
				\multicolumn{2}{|l|}{our framework (w, w)} & \multicolumn{2}{c|}{55.19 +/- 0.41}  & \multicolumn{2}{c|}{\textbf{64.95 +/- 1.11}}                                       & \multicolumn{2}{c|}{\textbf{40.88 +/- 0.63}}  \\ \hline
			\end{tabular}		
	}
	\caption{Average SF F1 score on 100 test episodes when $ U_{max} = 100 $.}
	\label{tab:1f1}
\end{table*}

\subsection{Datasets}
We conduct experiments on three benchmark datasets ATIS \cite{naacl/HemphillGD90}, SNIPS ~\cite{snips}, and TOP ~\cite{emnlp/GuptaSMKL18}. In pre-processing procedure, we follow \cite{handful-icsf} to modify slot label name by adding the associated intent label name as a prefix to each slot. 

We divide the dataset into train set (70\%), development set (15\%), and test set (15\%) respectively. For the SNIPS dataset, we choose not to form a development set. This is because that there are only 7 intents in the SNIPS dataset, and we require a minimum of 3 intents per split. Table \ref{tab:stat} provides the detailed dataset statistics.

\subsection{Baselines}
Following the work of Amazon AI~\cite{handful-icsf}, we compare our framework with some popular few-shot models: first order approximation of model agnostic meta learning (foMAML) \cite{finn2017model}), prototypical networks (Proto), and a fine-tuning method (Fine-tune) \cite{conf/naacl/GoyalMM18}. For each model, its embedding layer could be GloVe word embeddings (GloVe), GloVe word embeddings concatenated with ELMo embeddings (ELMo), or BERT embeddings (BERT). 

Furthermore, we can train the above models with two modes. One is to train and test the model on a single dataset, the other is to apply joint training approach to train the model on all the three datasets and test it on a single dataset. For example, SNIPS means we train and test the baseline on SNIPS dataset, and SNIPS (joint) means we train the baseline on all the three datasets but test it on SNIPS dataset.

The above baselines have been performed by \citet{handful-icsf}, we directly reuse their reported results. And as the second training mode is time consuming, we train our proposed model with the first mode.

In addition, we compare with the latest method Retriever \cite{naacl/Retriever}, which is a span-level retrieval method that learns similar contextualized representations for spans with the same label via a novel
batch-softmax objective.

We also evaluate our framework under three cases: our framework (o, o), our framework (w, o) and our framework (w, w), where our framework (o, o) represents $ \mathcal{L} = \mathcal{L}_{IC_{pn}} + \lambda \mathcal{L}_{SF_{pn}} $, our framework (w, o) represents $ \mathcal{L} = \mathcal{L}_{IC_{pn}} + \lambda \mathcal{L}_{SF_{pn}} + \gamma {\mathcal{L}_{IC_{scl}}} $, our framework (w, w) represents $ \mathcal{L} = \mathcal{L}_{IC_{pn}} + \lambda \mathcal{L}_{SF_{pn}} + \gamma {\mathcal{L}_{IC_{scl}}} + \delta{\mathcal{L}}_{SF_{scl}} $. Our framework (w, w) is the whole model.

\begin{figure*}[t]
	\centering
	\subfigure[Pic.1: Original]{
		\begin{minipage}[t]{0.32\linewidth}
			\centering
            \includegraphics[width=1.7in]{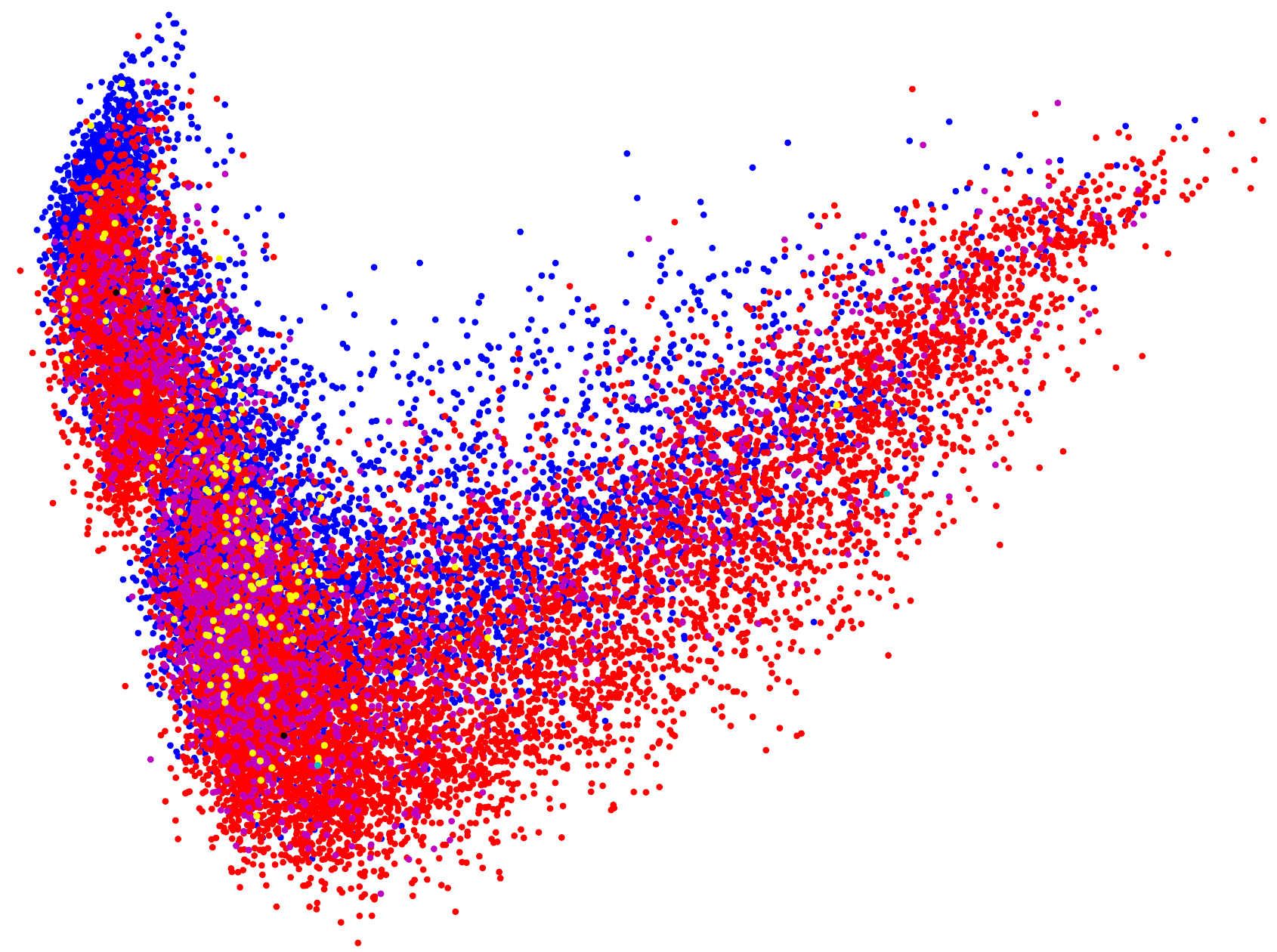}
		\end{minipage}%
	}
	\subfigure[Pic.2: CE]{
		\begin{minipage}[t]{0.32\linewidth}
			\centering
			\includegraphics[width=1.7in]{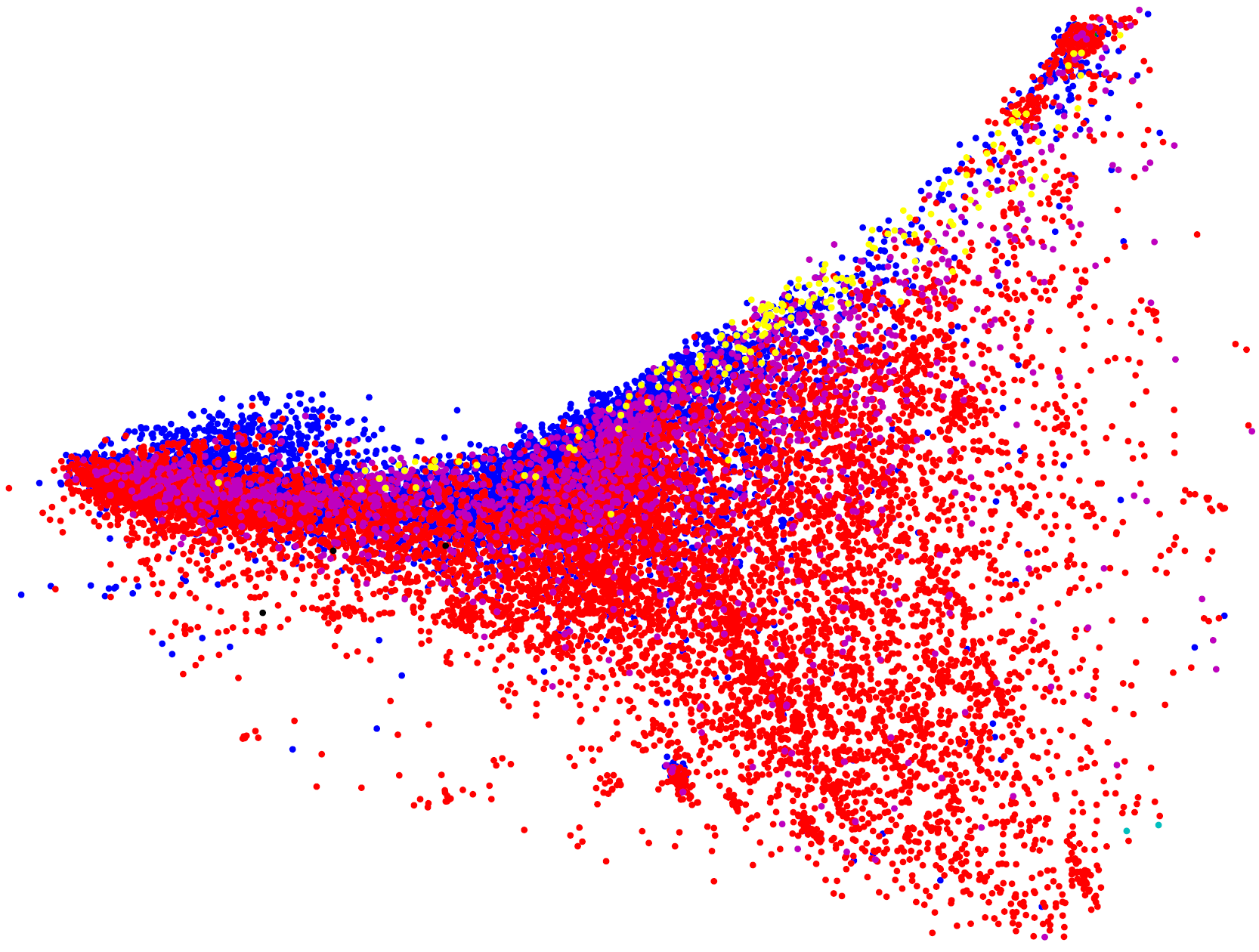}
		\end{minipage}%
	}
	\subfigure[Pic.3: CE+SCL]{
		\begin{minipage}[t]{0.32\linewidth}
			\centering
			\includegraphics[width=1.7in]{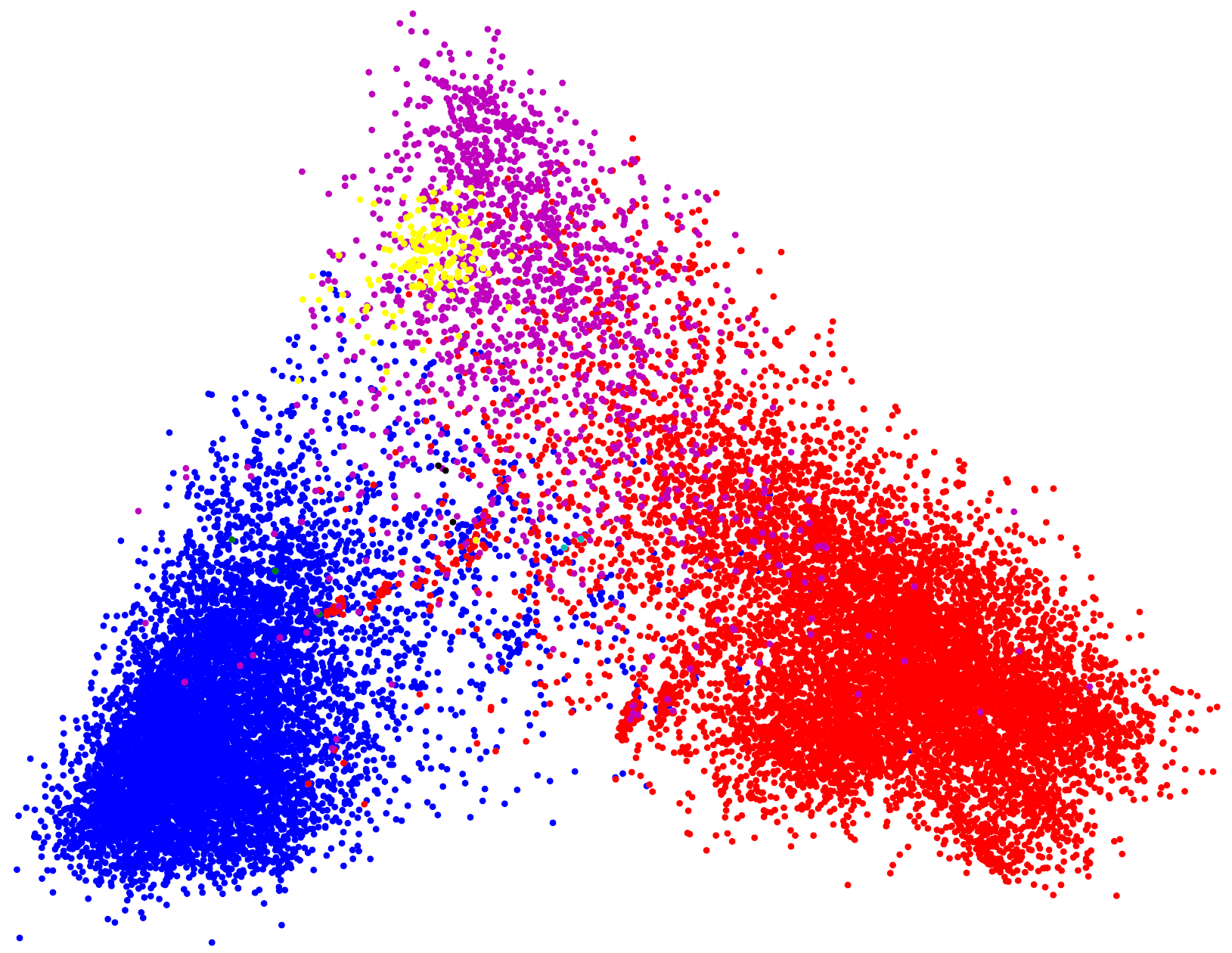}
		\end{minipage}
	}
	\centering
	\caption{Pic.1 shows sentence embeddings obtained from original pre-processing model without any training process. Pic.2 shows sentence embeddings via training the model with cross entropy (CE) loss of prototypical network. Pic.3 shows sentence embeddings via training the model with cross entropy (CE) and supervised contrastive loss (SCL). All the data are from TOP dataset. Data points with the same color come from the same class.}
	\label{Fig.2}
\end{figure*}

\subsection{Implementation Details}
\paragraph{Parameter Settings} In this paper, the dimension of hidden state is set to 1536 ($ d = 1536 $). We freeze 6 layers of BERT, and train all models using AdamW ~\cite{iclr/LoshchilovH19} optimizer with the initial learning rate $ 1 \times 10^{-4}$ and the dropout ratio 0.1. All the models are trained for 30 epochs. For hyperparameters $\lambda $, $ \gamma $ and $ \delta $, we use the grid searching method to determine them in the range (0, 1). For the hyperparameter $\tau$, we set $ \tau = 0.1 $ consistently.

\paragraph{Evaluation Metrics} We evaluate the performance of intent classification and slot filling with accuracy (Acc) and F1 score (F1), respectively.

\subsection{Result Analysis}
\paragraph{IC Performance}
Table \ref{tab:0acc} summarizes the average IC accuracy over 100 test episodes when the maximum support set size $ U_{max} = 20 $, where the top 2 results are highlighted in bold. We could make the following observations. (1) When comparing with the baselines that use the same word embeddings (BERT), our framework (w, w) improves upon the strong baseline BERT+Proto by nearly 4\%, 22\% and 10\% on SNIPS, ATIS and TOP respectively, which shows the superiority of our proposed model. (2) When comparing with all the baselines, our framework (w, w) improves upon the strong baseline ELMo+Proto by nearly 15\% and 12\% on ATIS and TOP respectively. (3) On SNIPS dataset, our framework (w, w) performs a little worse than ELMo+Fine-tune with joint training mode. This is because that ELMo+Fine-tune with joint training mode trains the model on all the three datasets, but our framework only trains the model on SNIPS. In addition, the word embeddings of ELMo seem more suitable for SNIPS.

Table \ref{tab:1acc} shows the average IC accuracy over 100 test episodes when the maximum support set size $ U_{max} = 100 $, where the top 2 results are highlighted in bold. We could make the similar observations. (1) Our framework (w, w) performs the best when comparing with the baselines that use the same word embeddings. (2) Except for SNIPS on which ELMo+Fine-tune and ELMo+Proto get the best two results, our framework (w, w) always performs better than other baselines.

\paragraph{SF Performance}
Table \ref{tab:0f1} and Table \ref{tab:1f1} summarize the average SF F1 score over 100 test episodes when the maximum support set size $ U_{max} = 20 $ and $ U_{max} = 100 $ respectively, where the top 2 results are highlighted in bold. It can be seen that (1) When comparing with the baselines that use the same word embeddings (BERT), our framework (w, w) performs the best on all the datasets. (2) When comparing with all the baselines, our framework (w, w) can also obtain satisfactory performance in most cases.

\begin{table}[t]
    \scriptsize
	\centering
	\setlength\tabcolsep{3pt}
	\begin{tabular}{|c|cc|cc|cc|}
		\hline
		\multicolumn{1}{|c|}{\multirow{2}{*}{Model}} & \multicolumn{2}{c|}{SNIPS} & \multicolumn{2}{c|}{ATIS} & \multicolumn{2}{c|}{TOP} \\ \cline{2-7}
		\multicolumn{1}{|c|}{}                       & \multicolumn{1}{c}{IC Acc} & \multicolumn{1}{c|}{SF F1} & \multicolumn{1}{c}{IC Acc} & \multicolumn{1}{c|}{SF F1} & \multicolumn{1}{c}{IC Acc} & \multicolumn{1}{c|}{SF F1} \\ \hline
		only intent-to-slot  & 81.20 & 49.57   & 72.90  &  59.27  & 57.11  & 36.12    \\
		only slot-to-intent  & 82.75 & 48.95   & 73.57  &  54.73  & 58.39  & 34.43    \\
		our framework (o, o)  & 84.61 & 50.03   & 76.09  &  61.79  & 59.63  & 38.41    \\ \hline
	\end{tabular}
	\caption{Ablation study on the ATIS, SNIPS and TOP datasets when $ U_{max} = 20$.}
	\label{tab:ablation}
\end{table}

\subsection{Ablation Study}
\paragraph{Explicit-Joint Learning}
To verify the effectiveness of slot-attention-based intent representation and intent-attention-based slot representation, we make the ablation study. The results when $ U_{max} = 20$ are shown in Table \ref{tab:ablation}. Our framework (o, o) is the model that only contains explicit-joint learning. Only slot-to-intent represents the model that only uses slot-attention-based intent representation while replacing intent-attention-based slot representation with pure slot representation. Similarly, we have the only intent-to-slot model. From the results, it can be seen that our framework (o, o) performs better than the other two baselines, which demonstrates the effectiveness of extracting intent and slot representations via bidirectional interaction.

\paragraph{Supervised Contrastive Learning}
Our proposed objective function includes a \textit{cross entropy (CE)} term of prototypical network and \textit{supervised contrastive learning (SCL)} term, the latter aims to push samples in the same class close and samples in different classes further apart. By comparing the results of our framework (w, o) with our framework (o, o) in Table \ref{tab:0acc} and Table \ref{tab:1acc}, we can get that the term $ \mathcal{L}_{IC_{scl}} $ brings nearly 0.1\% $\sim$ 4.3\% improvement for IC accuracy. By comparing the results of our framework (w, w) with our framework (w, o) in Table \ref{tab:0f1} and Table \ref{tab:1f1}, it can be seen that the term $ \mathcal{L}_{SF_{scl}} $ brings nearly 0.6\% $\sim$ 2.9\% improvement for SF F1 score. The performance improvement demonstrates the effectiveness of the SCL loss for both IC and SF tasks.

Figure \ref{Fig.2} visualizes the distribution of sentence embeddings in TOP dataset, we can observe that the original distribution is random in Pic.1. As shown in Pic.2, CE can separate the data in different classes to some extent. In Pic.3, SCL term further encourages more compact clustering of the data points in the same class.

\section{Conclusion}
In this paper, we propose a new and practicable framework for few-shot intent classification and slot filling. The performance gains of our method come from two aspects: explicit-joint learning and supervised-contrastive learning. By explicit-joint learning, we can effectively utilize the close relationship between IC and SF tasks. By supervised-contrastive learning, we can obtain more class-indicative representations. We thoroughly evaluate our framework on few-shot IC and SF tasks and achieve impressive performance on three public datasets SNIPS, ATIS and TOP. In future work, we plan to explore more explicit-joint learning strategies and extend our framework to deal with multiple-intent classification.

\section*{Acknowledgements}
The authors are grateful to the anonymous reviewers for their valuable comments and suggestions. This work was supported by National Natural Science Foundation of China (No. 62106035, 61876028), and the Fundamental Research Funds for the Central Universities (No. DUT20RC(3)040, No. DUT20RC(3)066).

\bibliography{anthology}
\bibliographystyle{acl_natbib}

\end{document}